# Hardware-in-the-Loop Simulation for Evaluating Communication Impacts on the Wireless-Network-Controlled Robots

Honghao Lv[1, 2]
lvhonghao@zju.edu.cn
honghaol@kth.se

Zhibo Pang[2, 3]
zhibo@kth.se
pang.zhibo@se.abb.com

Ming Xiao[2]
mingx@kth.se

Geng Yang[1]
yanggeng@zju.edu.cn

1. *School of Mechanical Engineering, Zhejiang University*
Hangzhou, China

2. *EECS, Department of Intelligent Systems, Royal Institute of Technology (KTH)*
Stockholm, Sweden

3. *ABB Corporate Research Sweden*
Västerås, Sweden

*Abstract*— **More and more robot automation applications have changed to wireless communication, and network performance has a growing impact on robotic systems. This study proposes a hardware-in-the-loop (HiL) simulation methodology for connecting the simulated robot platform to real network devices. This project seeks to provide robotic engineers and researchers with the capability to experiment without heavily modifying the original controller and get more realistic test results that correlate with actual network conditions. We deployed this HiL simulation system in two common cases for wireless-network-controlled robotic applications: (1) safe multi-robot coordination for mobile robots, and (2) human-motion-based teleoperation for manipulators. The HiL simulation system is deployed and tested under various network conditions in all circumstances. The experiment results are analyzed and compared with the previous simulation methods, demonstrating that the proposed HiL simulation methodology can identify a more reliable communication impact on robot systems.**

## I. INTRODUCTION

With the growing network technologies and the Industrial Internet of Things (IIoT) towards Industrial 4.0, Industrial Edge Computing enables the automation platform that can be directly accessed from the cloud [1]. This makes various wireless-network-controlled robot platforms practically applied to industrial production [2]. To ensure the robustness and adaptability of the robotic system, simulation and testing of various communication conditions are usually required during the deployment process and design stage of a robot platform in an industrial facility [3]. Introducing delays or other communication characteristics into the controller codes to imitate the impact of the network is a common solution for the simulation [4]. However, the practical communication condition, particularly the wireless network, is extremely difficult to be modeled and simulated [5]. Hardware-in-Loop (HiL) simulation refers to the simulation technology which simulates one part of the whole system with computer modeling while using physical modeling or an actual system for the other part [6].

In the past few years, many simulation methods have been developed for robot system design and optimization. In [7], a HiL simulation system for manipulator control is built to improve the system stability by deploying the self-designed PID controller. Lamping *et al.* worked on a multi-agent Unmanned Aerial Vehicles (UAV) system based on Robot Operation System (ROS), and they experimented with control and supervision algorithms on multiple UAVs with the designed HiL system [8]. Most of the above examples of the HiL simulation and many more about the robotic applications are aimed to provide a flexible and easy-to-use environment interface for a controller design. One common constraint of these HiL systems related to robotics, mechatronics, and control is the limited modeling of the communication condition, especially for the growing wireless-network-controlled robots recently [9]. This means simulated the network condition for the robot systems in the design stage is still a challenge for the robot developer.

In this study, we propose a potential HiL simulation methodology for helping to involve the influences of the physical network between the controller and robot model, as well as validating the communication impacts on the wireless-network-controlled robots. This work provides an easy-to-use and modularized interface for the robotic system quality engineer to conduct the simulation under real network

This work was supported in part by the Swedish Foundation for Strategic Research (SSF) through the project APR20-0023. The work of Honghao Lv and Geng Yang was supported in part by the National Natural Science Foundation of China under Grant 51975513 and Grant 51890884; in part by the Natural Science Foundation of Zhejiang Province under Grant LR20E050003; in part by the Major Research Plan of Ningbo Innovation 2025 under Grant 2020Z022; and in part by the Zhejiang University Special Scientific Research Fund for COVID-19 Prevention and Control under Grant 2020XGZX017. Honghao Lv gratefully acknowledges financial support from China Scholarship Council.

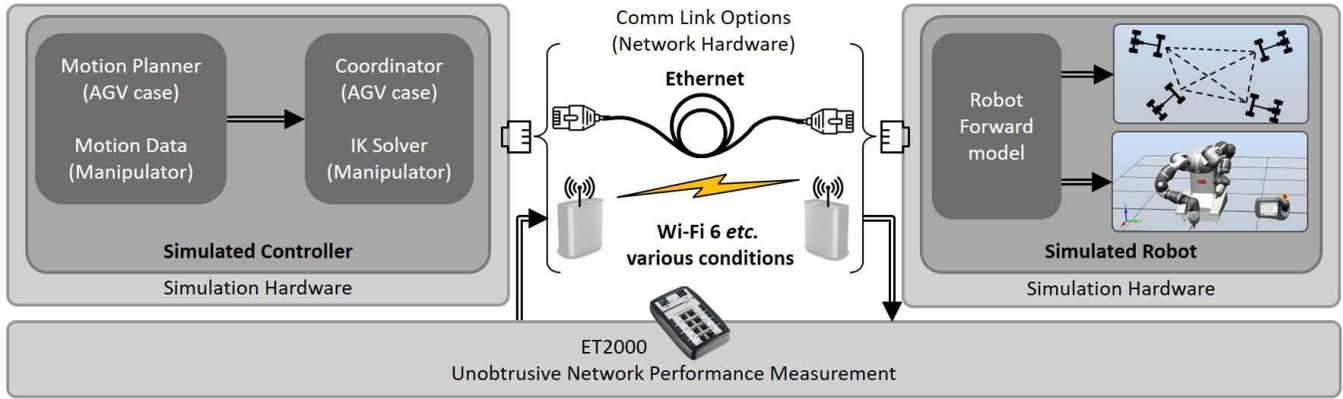

Fig. 1. Diagram of the network hardware-in-the-loop simulation system.

conditions, without having to model sophisticated communication or change the initial controller.

## II. HARDWARE-IN-THE-LOOP SIMULATION METHODOLOGY

### A. System Architecture

As shown in **Fig. 1**, the robot simulated platform is commonly separated into the controller part and the robot model part. The latter usually includes the robot kinematic forward model or the dynamic model for control simulation. Two robotic application case studies, that are considerably affected by network communication conditions, are chosen and deployed to verify the HiL simulation system in this work. The first is the safe multi-robot coordination case, in which the real-time motion status and the navigation command of each robot in the fleet require a reliable wireless communication condition to avoid collision with each other. The second one is a manipulator teleoperation case based on the captured human motion data. The high-frequency motion data transfer between the robot and the human must be guaranteed by reliable communication conditions. In the first case study, the motion planner and the coordinator for the multi-robot coordination belong to the simulated controller part. The robot forward model for every single mobile robot, including the robot size and motion parameters in the 2-D coordinate space, is the robot model part to be controlled. For the case study of the manipulator teleoperation, the human motion processor and the inverse kinematic solver are treated as the controller part, while the robot kinematic model in the 3-D coordinate space and visualization tool is the robot model part to be controlled. The selected two case studies is the respresentive applied cases in industrial robot area, and for other robot senerios, the proposed HiL simulation framework can be transfered and deployed using the native codes and models in the on-site controllers.

The hardware interface for the network devices is designed and injected between the controller and the robot model, providing practical communication options. Both the Ethernet and Wi-Fi 6 communication conditions can be chosen for the HiL simulation. Meanwhile, the unobtrusive network performance measurement using ET2000 is integrated into the system for parallel network monitoring. The user can measure the timestamp-based real-time network performance which corresponds to the real-time robot motion performance.

### B. Case Study 1: HiL Simulation Framework for Safe Multi-Robot Coordination

The first HiL simulation case study is based on the open-source Coordination_ORU project from Orebro University [10]. The Coordination_ORU platform provides a centralized coordination simulator in two-dimensional (2D) cartesian coordinate space, integrated with the motion planner for fleets of mobile robots. As shown in **Fig. 2**, the motion planner is used to compute the path and trajectory for each robot from the given start location to the target location. The coordinator is in charge of handling the multi-robot order in the critical section (overlap of multi-robot trajectory envelops), using the heuristic robot ordering policies, to avoid collisions and ensure mobile safety. The simulated robot forward model is driven by the motion command from the controller, and the real-time multi-robot motion status is displayed in the designed simulation graphical user interface (GUI), as shown in **Fig. 3**. The robot trajectory tracker is used to track the robot trajectories and send the real-time robot status back to the controller.

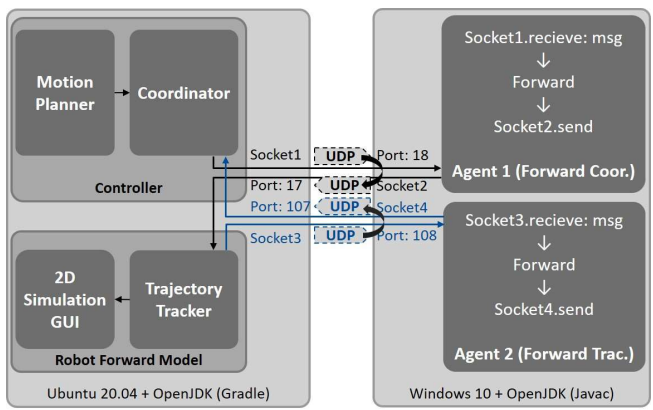

Fig. 2. HiL Simulation Software framework for safe multi-robot coordination.

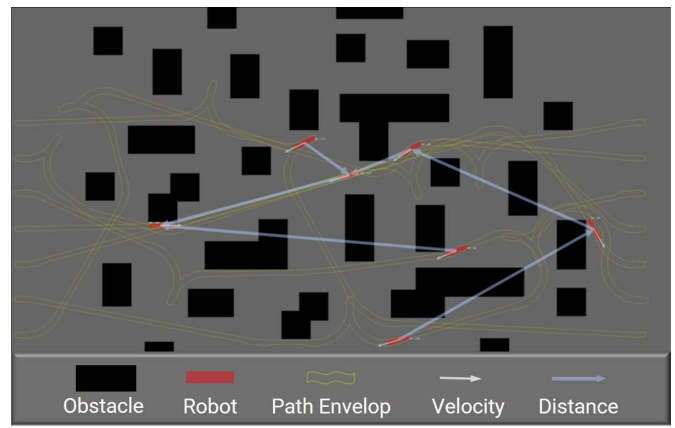

Fig. 3. GUI example of the simulated multi-robot coordination platform.

The controller and the robot forward model are achieved by Java and built by Gradle running on Ubuntu 20.04. To involve the real network condition, two agents are designed to forward the robot command and the robot status, as illustrated in **Fig. 2**. Agent 1 is responsible to forward the robot motion command from the controller to the simulated robot model, which is done over two independent UDP sockets. Agent 2 is responsible to forward the current robot status from the driven robot model to the controller. Two independent UDP sockets are used here as well. These two agents are programmed by java and deployed on another windows 10 computer. The real network devices can be deployed between the simulated robot system and the agent system to evaluate the Ethernet and Wi-Fi 6 conditions. The original Coodination_ORU project modeled the network latency, and packet loss rate (PLR), using a Bernoulli distribution, for unreliable communication simulation [11]. Even though, the proposed HiL simulation method provides a more reliable solution for system safety verification than the static network modeling method, which is verified by experiments.

### C. Case Study 2: HiL Simulation Framework for Manipulator Teleoperation

The manipulator teleoperation is chosen as the second case study for evaluating communication impacts on wireless-network-controlled robots. In this part, a modified human-robot motion transfer teleoperation simulated system from our previous work [12] is deployed. This system uses the captured human motion data to control the dual-arm collaborative robot YuMi. As shown in **Fig. 4**, the software Axis Neuron running on Windows 10 captures human motion and sends the data to the robot controller via ROS serial protocol from Windows 10 to Ubuntu 20.04. The controller is responsible for processing the human motion data and converting the human TF data to robot tool center point data using the designed mapping strategy [13]. The robot tool center point data is used to calculate the real-time joint configuration by the inverse kinematic (IK) solver. The motion controller program is connected to the motion server running on the simulated robot model via TCP protocol. The motion server is responsible to control the normal motion like the linear motion and the joint motion that uses the given robot target. To achieve real-time motion mapping between the human and robot, the External Guide Motion (EGM) interface is used here, which provides a low-level interface with high-frequency responsiveness for robot control.

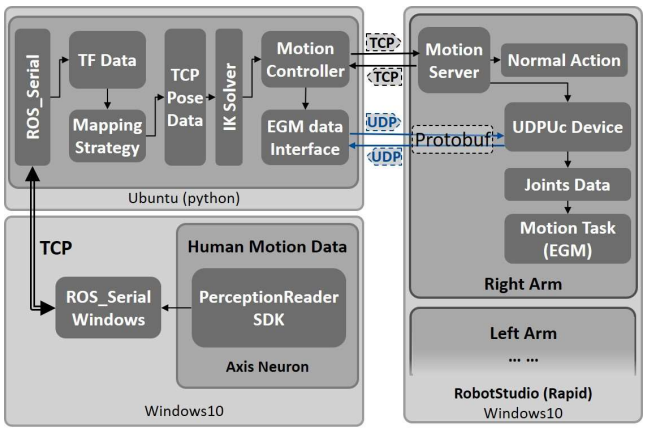

Fig. 4. HiL simulation software framework for manipulator teleoperation.

The UDP User Communication (UDPUc) device is defined for the task of each arm, which receives the joint values at a high rate. The EGM motion is activated/deactivated by the motion server. When the EGM motion is activated, the generated joint value queue will be sent using Google Protocol Buffers (*Protobuf*) based on the UDP protocol. For this case, the communication between the human motion capture part on the Windows 10 system and the ubuntu 20.04 is connected by Ethernet. The network hardware is injected between the controller running on ubuntu and the simulated YuMi robot in RobotStudio on the other Windows 10 system. Like the network condition setting in case study 1, both the Ethernet and Wi-Fi 6 wireless communication options can be chosen to test the physical communication impacts on the human-motion transfer teleoperation systems. The user can adjust the communication condition in the HiL simulation system according to their actual application conditions.

### D. Unobtrusive Network Performance Measurement

The ET2000 multi-channel probe is a versatile piece of hardware for analyzing the Industrial Ethernet solution. In this work, we integrated the ET2000 device into the HiL simulation system for monitoring the physical network performance used for each simulation case. The high timestamp resolution of 1 ns enables very precise timing analysis of the connected network segments, which reduces the influence on the original system very small. This work focused on the system framework achievement and verification of the HiL simulation. The related software design of the unobtrusive network performance measurement part is taken as a reference from the ABB internal reports, which will not be introduced in detail here.

## III. EXPERIMENTS AND SYSTEM EVALUATION

Corresponding experiments on the above two cases using the proposed HiL simulation framework are designed and carried out to verify the feasibility and practicality. For the safe multi-robot coordination case, the experiments both for physical Ethernet conditions and various wireless network conditions are implemented, compared with the static network modeling methodology. In the manipulator teleoperation case study, the experiments under Ethernet conditions are conducted as the baseline, then the experiments under the various real wireless network conditions are conducted using the HiL simulation system to evaluate the communication impacts on manipulator teleoperation based on human-robot motion transfer.

### A. Experiment Setup for the Wireless Network Condition

In this work, we chose the Wi-Fi 6 as the wireless condition, and a pair of ASUS AX6600 WiFi 6 routers are selected as the hardware. These two routers can be connected, providing a wireless connection based on Wi-Fi 6 to replace the wired network. One router is selected as the master and the second is the slave router. Expected to the Ethernet wired network test, the wireless network condition is designed as below:

*1) Short-range Wi-Fi 6 connection:* As shown in **Fig. 5**, the master router is placed in the Lab Area 1 which is connected to the simulated robot platform for multi-robot coordination and the simulated controller for the manipulator teleoperation. For the short-range wireless network test, the slave master is placed in the Lab Area 2, next to the Lab Area 1 at a distance of 20 m to the master router.

*2) Long-range Wi-Fi 6 connection:* For the long-range wireless connection test, the master router is placed in the same location in the Lab Area 1. As shown in **Fig. 5**, the slave router is placed at the end of the corridor next to the Fika Room, with a distance of around 35 m to the master router. The salve router is connected to the Agent software for the multi-robot coordination and the simulated YuMi robot for the manipulator teleoperation case.

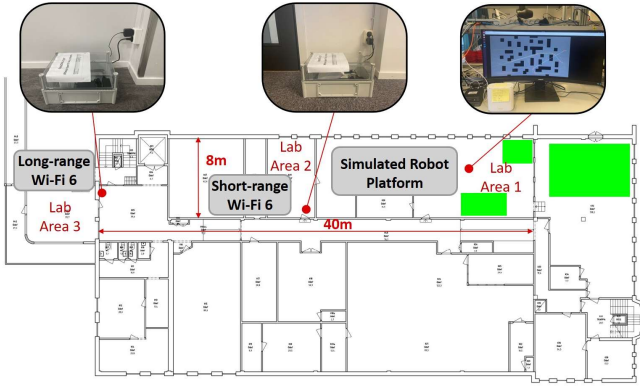

Fig. 5. Wireless network condition setup for HiL simulation experiments.

### B. HiL Simulation Result for Case Study 1: Safe Multi-Robot Coordination

This part is to verify the performance of the proposed HiL simulation system for the safe multi-robot coordination test. The parameters of two practical Automated Guided Vehicles (AGV) use cases in the industry application scenarios, harbor and warehouse, are chosen for the experiments. Harbor AGV is with a bigger size and higher moving speed [14], while the warehouse is with a smaller size and lower moving speed [15]. The parameters set for the simulation are listed in Table I. Seven robots are simulated and the static map with serval obstacles is set for each case as shown in **Fig. 3**.

TABLE I. MULTI-ROBOT COORDINATION SIMULATION PARAMETERS

| Application Case | AGV Parameters | | | |
|---|---|---|---|---|
| | Number of robots | Single robot size[a] (m) | Max vel. (m/s) | Max acc. (m/s²) |
| Harbor AGV | 7 | 14.8 × 3.0 | 6 | 2 |
| Warehouse AGV | 7 | 2 × 0.5 | 2 | 1 |

The static modeling methodology was conducted first, modeling the communication channel with various sets of PLR and the packet delay for the message transfer on one control cycle. Then the HiL simulation was implemented with the network condition of Ethernet, short-range Wi-Fi 6, and long-range Wi-Fi 6. During the simulation, each case was tested for at least 10000 critical sections (CS) created in the coordination process, and all the collisions were counted for each case. The PLR was set at 0 and 0.1, and the packet delay was set at 0, 10, 50, and 100 ms.

TABLE II. EXPERIMENTS RESULTS AND COMPARISON FOR HARBOR AGV MULTI-ROBOT COORDINATION CASE

| Harbor AGV Setup | | | Statistical data | | |
|---|---|---|---|---|---|
| Simulation methods | PLR | Delay (ms) | Total CS counts | Collison counts | $P_{collision}$ (×10³) |
| Static Modeling | 0 | 0 | 10018 | 0 | 0 |
| | | 10 | 10001 | 0 | 0 |
| | | 50 | 10003 | 0 | 0 |
| | | 100 | 10018 | 5 | 0.499102 |
| | 0.1 | 0 | 10003 | 0 | 0 |
| | | 10 | 10007 | 0 | 0 |
| | | 50 | 10092 | 0 | 0 |
| | | 100 | 10061 | 18 | 1.789087 |
| HiL | Ethernet | | 10819 | 0 | 0 |
| | Short-range Wi-Fi 6 | | 15489 | 14 | 0.903867 |
| | Long-range Wi-Fi 6 | | 10004 | 23 | 2.299080 |

The experiments of the multi-robot coordination simulation for the Harbor AGV case and the Warehouse case are listed in **Table II** and **Table III** separately. From the results of the experiment, using the static modeling methodology, there was no collision in the 10000 critical sections test under the 50 ms delay set for any case. For the bigger and faster Harbor AGV case, some collisions were raised when the delay is set as 100 ms. The probability of collision $P_{collision}$ is 0.499 ‰ for zero PLR and 100 ms delay. The $P_{collision}$ is 1.789 ‰ for 0.1 PLR and 100 ms delay. Then the HiL simulation results for the Harbor AGV case are listed below in **Table II**. The $P_{collision}$ is 0.904 ‰ and 2.299 ‰ respectively under the short-range and the long-range Wi-Fi 6 conditions. For the smaller and slower Warehouse AGV case, no collision happened when using the static modeling methodology. When using the HiL simulation method, the $P_{collision}$ is 0.099 ‰, 1.477 ‰, and 2.300 ‰ respectively under the Ethernet, short-range, and long-range Wi-Fi 6 conditions.

TABLE III. EXPERIMENTS RESULTS AND COMPARISON FOR WAREHOUSE AGV MULTI-ROBOT COORDINATION CASE

| Warehouse AGV Setup | | | Statistical data | | |
|---|---|---|---|---|---|
| Simulation methods | PLR | Delay (ms) | Total CS counts | Collison counts | $P_{collision}$ |
| Static Modeling | 0 | 0 | 13859 | 0 | 0 |
| | | 10 | 10001 | 0 | 0 |
| | | 50 | 10001 | 0 | 0 |
| | | 100 | 10015 | 0 | 0 |
| | 0.1 | 0 | 10002 | 0 | 0 |
| | | 10 | 10001 | 0 | 0 |
| | | 50 | 10037 | 0 | 0 |
| | | 100 | 21645 | 0 | 0 |
| HiL | Ethernet | | 10004 | 1 | 0.099960 |

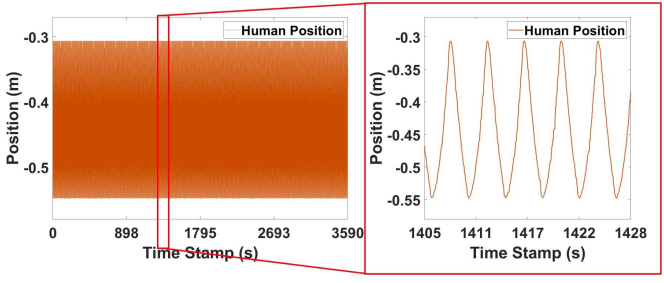

Fig. 6. Input human motion data for manipulator teleoperation.

| Warehouse AGV Setup | | | Statistical data | | |
|---|---|---|---|---|---|
| **Simulation methods** | **PLR** | **Delay (ms)** | **Total CS counts** | **Collison counts** | **$P_{collision}$** |
| Short-range Wi-Fi 6 | | | 10155 | 15 | 1.477105 |
| Long-range Wi-Fi 6 | | | 17394 | 40 | 2.299644 |

For most Wi-Fi network condition, the delay under 50 ms is considered meaningful. So HiL simulation results get more high results of $P_{collision}$ compared to the common Wi-Fi modeling, which is a more realistic simulation under the actual network conditions for the multi-robot coordination system. When we model the communication network using the common KPIs, even under probability distribution model, some uncommon factors for the network, such as congestion or the sync problem special for certain control cases cannot be reproduced.

### C. HiL Simulation Result for Case Study 2: Manipulator Teleoperation

In order to show the communication impacts on the manipulator teleoperation based on human-robot motion transfer, we conducted case study 2 under the real network conditions of Ethernet, short-range Wi-Fi 6, and long-range Wi-Fi 6 using the proposed HiL simulation platform. For the human motion input side, the motion data is sent under the frame frequency of 125 Hz. The human swung his right arm from one side to another side, and the motion was recorded and repeated play for at least 60 minutes. Each motion loop took 4.033s, so a total of 890 motion loops were played for one test case. According to the motion mapping strategy we deployed here, the mapping scale was set as 0.8 between the human motion and the robot motion. The played human motion data are plotted as **Fig. 6**. In the simulated robot side in RobotStudio, we deployed a smart component with the tool center pose sensing function and logged the real-time robot motion data. The robot tool center point motion data in the y-axis direction under the Ethernet condition is plotted in **Fig. 7**. And the motion data in the y-axis direction under the short-range and long-range Wi-Fi 6 condition are shown respectively in **Fig. 8** and **Fig. 9**.

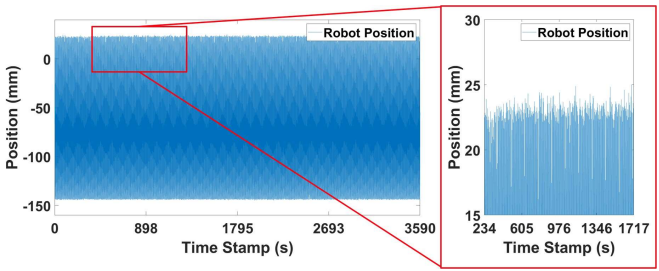

Fig. 7. Simulation result: measured robot position in Ethernet communication condition.

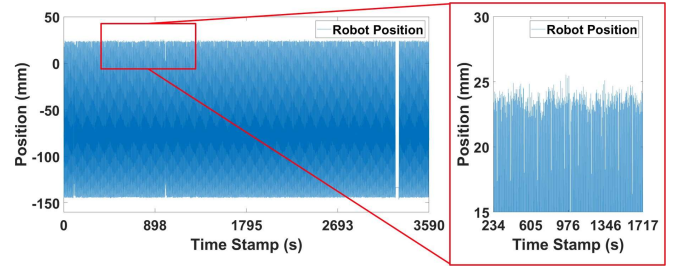

Fig. 8. Simulation result: measured robot position in short-range Wi-Fi 6 communication condition.

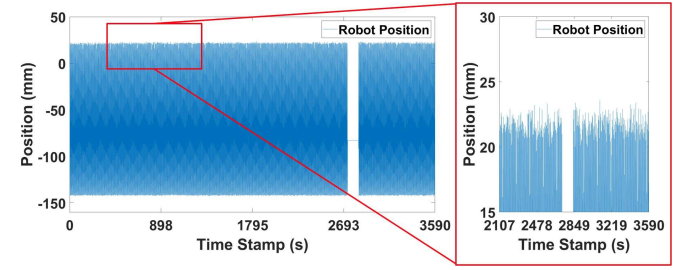

Fig.9. Simulation result: measured robot position in long-range Wi-Fi 6 communication condition.

From the experiment results of the robot motion, with the network condition changed from Ethernet to Wi-Fi 6, the measured robot motion performance changed obviously. Compared with the Ethernet condition, some loss of the motion loop can be observed under the wireless network condition. To eliminate errors and delay introduced by computation, we plot the comparison between the desired joint position sent from the controller side and the measured real joint position from the simulated robot side, as shown in **Fig. 10**. And the Joint position error between the desired joint position and the measured joint position under the short-range Wi-Fi 6 condition is shown in **Fig. 11**. In order to quantify the impact of communication on robot motion during the manipulator teleoperation case, we defined the motion loss rate $MLR$ by (1) as below:

$$MLR = \frac{N_S - N_a}{N_S} \tag{1}$$

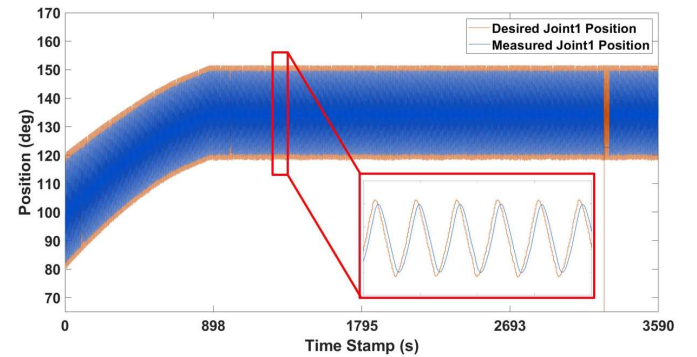

Fig. 10. Comparison of the desired joint position and the measured real joint position under short-range Wi-Fi 6 condition.

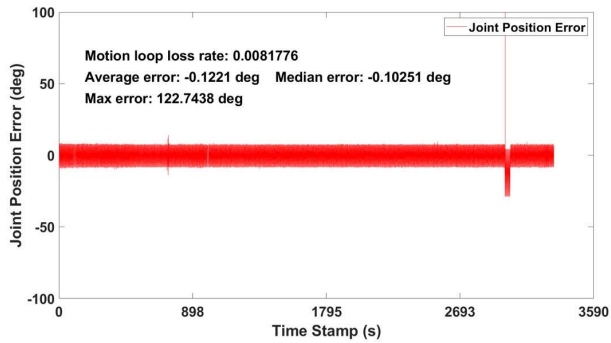

Fig. 11. Joint position error under short-range Wi-Fi 6 communication condition.

where the $N_s$ is the counts of the motion loop sent from the controller and the $N_a$ is the counts of the motion loop executed by the robot. All the motion loops are counted by the number of peaks of the motion data.

As a result, the *MLR* under the Ethernet condition is 0, while the *MLR* under the short-range and the long-range Wi-Fi 6 condition is 0.008 and 0.03 respectively. It can be seen that as the network conditions get worse, the *MLR* increases gradually. This is due to the instability of the network connection, which leads to an increase in the UDP data PLR or even the connection interruption of the EGM interface for teleoperation. The experiment results of the robot motion e under the wireless network conditions compared to the Ethernet condition shows the feasibility and practicability of the proposed HiL simulation methodology for the manipulator teleoperation simulation case.

## IV. CONCLUSION

This paper proposed an HiL simulation methodology to assess the impacts of communication on wireless-network-controlled robots. Two robot application cases, safe multi-robot coordination and manipulator teleoperation, are deployed and tested using the proposed HiL simulation methodology. And the simulation experiments under the Ethernet network condition and two wireless network conditions are conducted. Compared to the static modeling methodology for the multi-robot coordination case, the proposed HiL simulation methodology shows more realistic simulation results under the actual network conditions. Also, the communication impacts on the manipulator teleoperation case are investigated using the HiL simulation, which shows the practicability of the proposed methodology. The unobtrusive network performance measure interface is designed for future parallel testing and analysis of the network performance. Next, we will improve the software of the unobtrusive network performance measure interface, and conduct further real-time performance network testing synchronized with the simulated robot system.

## ACKNOWLEDGMENT

Z. Pang and G. Yang provided directional guidance to the research. H. Lv and Z. Pang conceived, designed, and performed the experiments. H. Lv, M, Xiao and G. Yang revised the paper.

The authors would like to express their gratitude to all the participants of this experiment.